%% file: main.tex
\pdfoutput=1

\documentclass[11pt]{article}

\usepackage{acl}

\usepackage{times}
\usepackage{latexsym}

\usepackage[T1]{fontenc}

\usepackage[utf8]{inputenc}

\usepackage{microtype}

\usepackage{times}
\usepackage{latexsym}

\usepackage{microtype}
\usepackage{hyperref}

\usepackage{graphicx}
\usepackage{xspace}
\usepackage{paralist}
\usepackage{booktabs}
\usepackage{multirow}
\usepackage{amsmath}
\usepackage{amsfonts}
\usepackage{amssymb}
\usepackage{pifont}
\usepackage{mathrsfs}
\usepackage{algorithm,algpseudocode}
\usepackage{mdwlist}
\usepackage{enumitem}
\usepackage{color}
\usepackage{dsfont}
\usepackage{amsthm}
\usepackage{bm}
\usepackage{float}
\usepackage{caption}
\usepackage{subfigure}

\newcommand{\loss}{\ensuremath{\mathcal{L}}}
\input{defs}

\newcommand{\method}{ConST\xspace}

\title{Cross-modal Contrastive Learning for Speech Translation}

\author{Rong Ye \textsuperscript{\rm 1}, 
    ~Mingxuan Wang \textsuperscript{\rm 1}, 
    ~Lei Li \textsuperscript{\rm 2}\thanks{~~Partial work was done while at ByteDance.} \\
    \textsuperscript{\rm 1} ByteDance AI Lab, 
    ~\textsuperscript{\rm 2} University of California, Santa Barbara \\
    {\tt \{yerong, wangmingxuan.89\}@bytedance.com} \\
    {\tt leili@cs.ucsb.edu}
}

\begin{document}
\maketitle

\begin{abstract}
\input{000_abstract}

\end{abstract}

\section{Introduction}
\input{010_intro}
\begin{figure*}[t]
    \setlength{\abovecaptionskip}{-0.1cm}
    \setlength{\belowcaptionskip}{-0.5cm}
    \centering
    \includegraphics[width=0.85\linewidth]{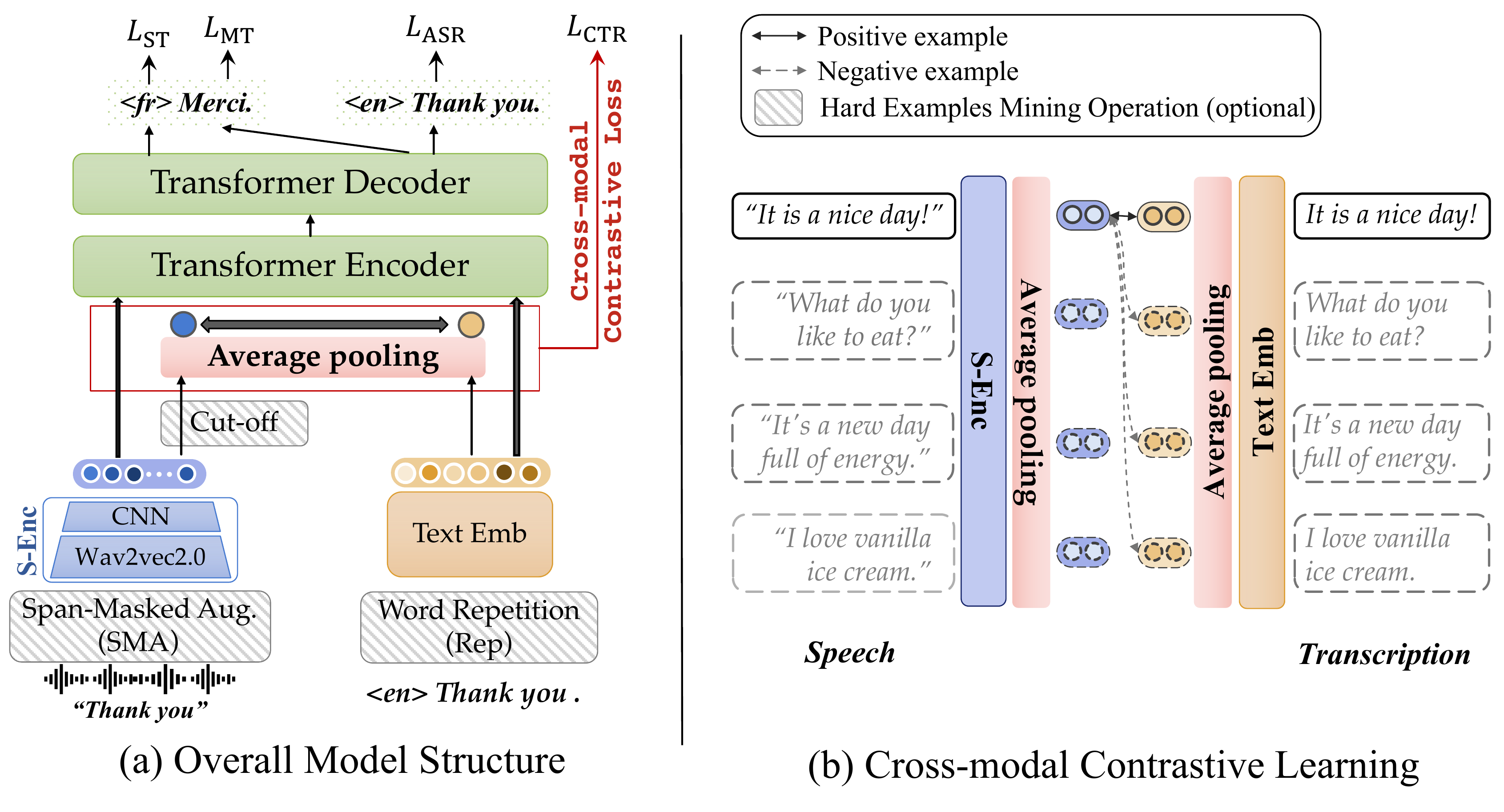}
    \caption{\textbf{Left}: Model structure of \method. The gray shaded modules are the \textit{optional} data augmentation operations introduced in Section~\ref{sec:method_DA}. \textbf{Right}: An illustration of cross-modal contrastive learning.}
    \label{fig:framework}
\end{figure*}

\section{Related Work}
\input{060_related}

\section{The \method Approach}
\input{020_method}

\section{Experiments}
\input{030_exp}

\section{Analysis}

\input{040_analysis}

\input{table_case_study}

\section{Why does cross-modal contrastive learning work? --- Analysis on the Modality Gap}
\input{050_analysis_gap}

\section{Case Analysis}
\input{093_case_study}

\section{Conclusion}
\input{070_conclusion}

\section{Broader Impact}
\input{080_broader_impact}

\bibliography{anthology,custom}
\bibliographystyle{acl_natbib}

\clearpage
\appendix
\input{090_appendix}

\end{document}

%% file: defs.tex
\newcommand{\notes}[1]{}

\theoremstyle{definition}

\theoremstyle{plain}

\newcommand{\veca}{\ensuremath{\mathbf{a}}\xspace}

\newcommand{\vecs}{\ensuremath{\mathbf{s}}\xspace}

\newcommand{\ith}[1]{\ensuremath{i^{{th}}}}

\newcount\permx
\newcount\permy
\def\permdot#1#2{
\permx=#1 \advance\permx by-1
\permy=#2 \advance\permy by-1
\psframe[fillcolor=black, fillstyle=solid]
(\permx,\permy)(#1, #2)
}

\newcommand\union{\cup}

\newcommand{\boxnum}[1]{{\setlength{\fboxsep}{1pt}\raisebox{1pt}{\hspace{1pt}\fbox{\tiny #1}\hspace{1pt}}}}
\newcommand{\ind}[1]{\ensuremath{_{\kern-0.5pt\boxnum{#1}}}}

\newcommand{\vecx}{\mathbf{x}\xspace}
\newcommand{\vecy}{\mathbf{y}\xspace}

\newcommand{\smallnt}[1]{\ensuremath{_{\mbox{\tiny PP}}}\xspace}

\newcommand{\pseudocode}{Algorithm}
\floatname{algorithm}{\pseudocode}

\iffalse

\else

\fi



%% file: 000_abstract.tex


How can we learn unified representations for spoken utterances and their written text?
Learning similar representations for semantically similar speech and text is important for speech translation.
To this end, we propose \textbf{\method}, 
a cross-modal contrastive learning method for end-to-end speech-to-text translation. 
We evaluate \method and a variety of previous baselines on a popular benchmark MuST-C.
Experiments show that the proposed \method consistently outperforms the previous methods on, and achieves an average BLEU of 29.4.
The analysis further verifies that \method indeed closes the representation gap of different modalities --- its learned representation improves the accuracy of cross-modal speech-text retrieval from 4\% to 88\%.
Code and models are available at \url{https://github.com/ReneeYe/ConST}.

%% file: 010_intro.tex
End-to-end speech-to-text translation (E2E ST) becomes important in many internet products and real applications. 
An E2E ST system accepts audio signals as the input and generates the target translation using a single model.  
Compared with the conventional cascade ST models, E2E ST models have achieved almost comparable~\cite{bentivogli2021cascade} or even superior~\cite{ansari2020findings, potapczyk2020srpol,xu2021stacked} performance.

The performance of an E2E ST model is still restricted for languages with relatively small parallel data, compared to text machine translation (MT).
Existing approaches for ST focus on using additional data from MT and automatic speech recognition (ASR).
This can be realized through pre-training approaches~\cite{zheng2021fused,dong2021listen,dong2021consecutive} or multi-task training frameworks~\cite{tang2021general,ye2021end,han2021learning}. 

\begin{figure}
    \setlength{\belowcaptionskip}{-0.5cm}
    \centering
    \includegraphics[width=\linewidth]{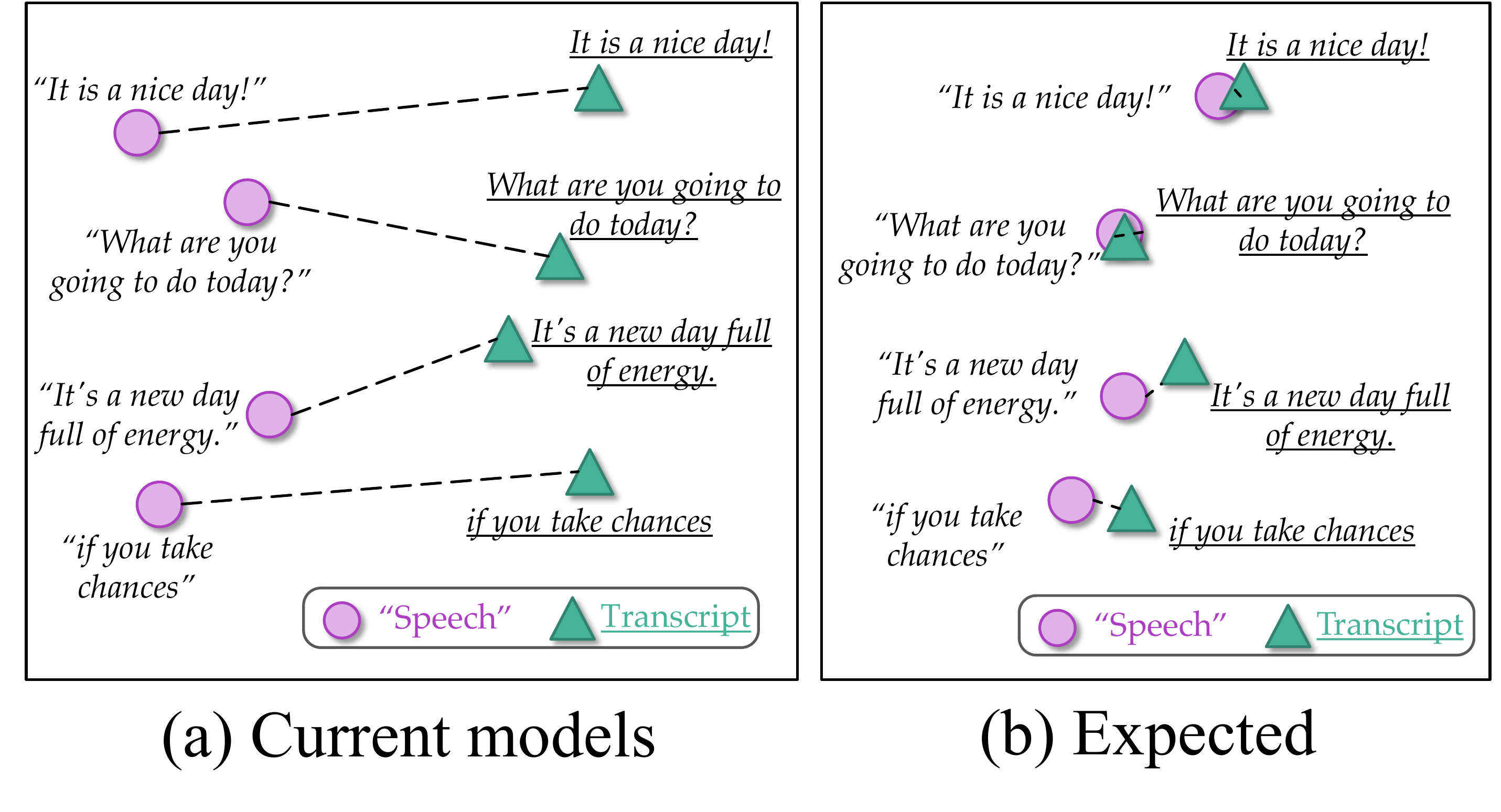}
    \caption{ Illustration of representations for speech and transcript text (projected to 2D). (a) representations learned by existing models. Pairs of speech and text representations are distant. (b) an ideal representation that we expect, where different modalities with same meaning should stay close to each other.}
    \label{fig:intro_strep}
\end{figure}

Different from the data perspective, this paper investigates the bottleneck of E2E ST from the neural representation perspective. 
We believe that when the representation of audio input is similar to its corresponding textual representation, it is easier for information to transfer from MT to ST, thus improving speech translation performance.

We analyze Transformer models for speech translation and observe a noticeable modality gap between encoder representations of speech and text (Sec.~\ref{sec:ana_gap} has more details) from existing ST models.
An ideal representation should satisfy: if the content of the speech and transcription are similar,  their encoded representations should likewise be close to each other.
Nevertheless, how to learn unified and aligned speech-text representations? 

Inspired by the recent progress of contrastive learning approaches in cross-lingual~\cite{lample2019cross,pan2021contrastive} and cross-modal vision-and-language domains~\citep{li2021unimo, zhou2020unified, dong2019unified}, we designed a simple \textbf{con}trastive learning method for \textbf{ST} (\textbf{\method}) to learn the representations that meet the aforementioned conditions explicitly.
On the one hand, our model inherits the advantages of the previous multi-task learning methods. 
On the other hand, it reduces the gap between the representations of speech and its corresponding transcription.

Our contributions are as follows.
\begin{itemize}[itemsep=1pt, leftmargin=10pt, parsep=0pt, topsep=1pt]
    \item We develop \method for speech translation, a cross-modal contrastive learning method, on top of the multi-task training framework.
    \item Our experiments on the MuST-C benchmark to show \method  achieves an average BLEU score of 29.4, outperforming the best previous baseline. 
    \item We show that \method indeed learns similar representations for two modalities and better retrieves text with speech input.
\end{itemize}

%% file: 060_related.tex
\noindent\textbf{End-to-end ST}~
To alleviate the error propagation in the cascaded ST systems and to make the deployment simpler, \citet{berard2016listen, weiss2017sequence} proposed to use an end-to-end architecture to directly translate speech into text in another language, without the intermediate transcription. 
\citet{kano2017structured,berard2018end,inaguma2020espnet, wang2020fairseqs2t,zhao2021neurst} implemented several off-the-shelf encoder-decoder E2E-ST models, such as BiLSTM~\cite{greff2016lstm} and Speech-Transformer~\cite{dong2018speech}.
However, training an end-to-end speech translation model is difficult because we need to design a cross-modal cross-language model, meanwhile, the \textit{speech-transcription-translation} supervised data for speech translation is significantly less than that of MT and ASR.
Methods, like data augmentation~\cite{park2019specaugmenta,pino2020self,chen2021specrec}, pre-training~\cite{weiss2017sequence,berard2018end,bansal2019pre, wang2020curriculum,alinejad2020effectively,dong2021consecutive,zheng2021fused}, self-training~\cite{pino2020self, wang2021large}, utilizing self-supervised pre-trained audio representation~\cite{wu2020self, han2021learning, ye2021end, wang2021large}, are proved to be effective.
Meanwhile, some work has shown that the encoder-decoder model with a single encoder cannot encode speech information well.
For example, \citet{dong2021listen} first proposed a second encoder to further extract semantic information of the speech sequence. \citet{xu2021stacked} proposed a stacked acoustic-and-textual encoder and introduced large-scale out-of-domain data. Also, multi-task frameworks ~\cite{le2020dual,tang2021general,ye2021end} are widely applied to further enhance the robustness for ST.
As a cross-modal task, some work has noted the problem of the modality gap. \cite{han2021learning} designed a fix-size semantic memory module to bridge such a gap, from the neuroscience perspective. However, we find that this approach actually sacrifices the effect of MT. So in this paper, we propose a simple yet effective contrastive learning method to bridge the gap and to improve ST performance. 

\noindent\textbf{Cross-modal grounding learning}~
This paper attempts to address the problem in speech translation from the perspective of cross-speech-text representation learning.
We are also inspired by cross-modal representation learning in the acoustic word embedding (AWE)~\cite{palaskar2019learned,kamper2020multilingual, hu2020multilingual} and the visual-language pre-training (VLP)~\cite{wu2019unified, lu2019vilbert, chen2020uniter, li2021unimo} tasks.
These works usually focus on enhancing textual representations with acoustic or visual information, in other words, grounding learning.
In this work, we consider the its dual form, i.e., grounding speech representations using text.

\noindent\textbf{Contrastive learning}~
Our method is motivated by the recent success in contrastive representation learning.
The contrastive learning method was first proposed to learn representations from unlabeled datasets (hence the term, self-supervised learning) by telling which data points are similar or distinct, especially in the field of computer vision~\cite{chopra2005learning, gutmann2010noise, schroff2015facenet, sohn2016improved, oord2018representation,chen2020simple, grill2020bootstrap}.
\citet{khosla2020supervised} extended the self-supervised batch contrastive approach to the fully-supervised setting and proposed a supervised contrastive learning method.
In speech processing, representative methods focused on speaker identification~\cite{ravanelli2018learning}, speech recognition~\cite{schneider2019wav2vec}, and audio representation learning~\cite{oord2018parallel,baevski2020wav2vec}. 
In the NLP area, the contrastive framework is used for sentence representation learning~\cite{fang2020cert, shen2020simple, gao2021simcse,wu2021esimcse, yan2021consert}, machine translation~\cite{pan2021contrastive}, and summarization~\cite{wang2021contrastive, cao2021cliff}. 
Very recently, contrastive learning is also applied to learning a unified representation of image and text~\cite{dong2019unified, zhou2020unified, li2021unimo}.
Motivated by the contrastive learning frameworks in cross-lingual and cross-modal topics, we introduce a similar idea in speech translation.

%% file: 020_method.tex
An end-to-end speech translation model directly translates audio sequence $\vecs=(s_1,...,s_{|\vecs|})$ to the text $\vecy=(y_1,...,y_{|\vecy|})$ in the target language.
Speech translation corpus $\mathcal{D} = \{(\vecs, \vecx, \vecy)\}$ provides \textit{transcript} $\vecx=(x_1,...,x_{|\vecx|})$ in the source language, as well.

In this section, we present the overall speech translation model and cross-modal contrastive learning. 
We also provide several feasible strategies to construct more positive and negative pairs to enhance the contrastive learning.

\subsection{Model Framework}
\label{sec:method_model}

\input{021_xstnet}

\input{022_train}

\subsection{Cross-modal Contrastive Learning}
\label{sec:method_CTR}
\input{023_crossmodal_contrastive}

\subsection{Mining Hard Examples for Contrastive Learning}
\label{sec:method_DA}
\input{024_data_aug}

%% file: 021_xstnet.tex
Our model consists four sub-modules: a speech encoder, a word embedding layer, a Transformer Encoder and a Transformer decoder (Figure~\ref{fig:framework}).
It is designed to take either speech or a sentence as input, and to output either source transcript or target translation text. Such architecture enables a universal framework for multiple tasks, including ST, MT and ASR. 

The \textit{speech encoder} module (S-Enc) is designed to extract low-level features for speech signals.
It contains Wav2vec2.0~\cite{baevski2020wav2vec} and two additional convolutional layers.
The input is raw waveform signal sampled at 16kHz. Each convolutional layer has a stride of 4 and $d$ channels. In total, it shrinks the time dimension by a factor of $4$.
Denote $\veca = \text{S-Enc}(\vecs)$ as the audio representation of the speech, $|\veca| \ll |\vecs|$.

Parallel to the speech encoder is the \textit{word embeeding layer}. It is the same as word embedding for text translation.

Both the speech encoder and word embedding layer are connect to \textit{Transformer encoder} and then passed to the \textit{Transformer decoder}. The Transformer encoder and decoder are using the same configuration as the original~\cite{vaswani2017attention}.
To explain, the \textit{Transformer encoder} further extracts the high-level semantic hidden representation of two modalities. The \textit{Transformer decoder} generates the word sequences (transcription and translation) for ST, MT and ASR tasks.
Since our model has a complete Transformer encoder-decoder as a sub-module, this makes it possible to pre-train using large-scale extra MT parallel data.

%% file: 022_train.tex
Previous work has shown that multi-task learning on ST, MT and ASR improves translation performance~\cite{indurthi2020data, tang2021general, ye2021end}.
Our training loss consists of the following elements.
\begin{equation}
    \loss =  \loss_{\text{ST}} + \loss_{\text{ASR}} + \loss_{\text{MT}} + \lambda \loss_{\text{CTR}}
    \label{eq:total_loss}
\end{equation}
where
\begin{align*}
    \loss_{\text{ST}} & = - \sum_{n} \log P(\vecy_n | \vecs_n) \\
    \loss_{\text{ASR}} & = - \sum_{n} \log P(\vecx_n | \vecs_n) \\
    \loss_{\text{MT}} & = - \sum_{n} \log P(\vecy_n | \vecx_n)
\end{align*}

The first three elements are cross-entropy losses on \textit{<speech, target text>}, \textit{<speech, source text>} and \textit{<source text, target text>} pairs. These pairs are built from the triplet ST data.
We also introduce a cross-modal contrastive loss term $\loss_{\text{CTR}}$ (see Section~\ref{sec:method_CTR} for details). It aims to bring the representation between the speech and textual transcription modalities closer (its effect will be analyzed in detail in Section~\ref{sec:ana_gap}). $\lambda$ is a tuned hyper-parameter of the weighted contrastive loss term.

%% file: 023_crossmodal_contrastive.tex
As mentioned in the beginning, since we need to produce similar representations for the speech and transcript sharing the same semantic meanings, we propose cross-modal contrastive learning method to bring their representations closer together.
The main idea of cross-modal contrastive learning is to introduce a loss that brings speech and its corresponding transcript (positive example) near together while pushing irrelevant ones (negative examples) far apart.

Given a positive example of such a speech-transcript pair $(\vecs, \vecx)$, we randomly pick a set of $N-1$ transcripts $\{\vecx_i^-\}_{i=1}^{N-1}$ from the same batch as negative examples.
For speech $\vecs$ and its transcript $\vecx$, we first average them in terms of the time dimension, 
\begin{align}
    u =\text{MeanPool}(\text{S-Enc}(\vecs)) \label{eq:audio_emb}\\
    v =\text{MeanPool}(\text{Emb}(\vecx)) \label{eq:text_emb}
\end{align}
and apply the multi-class N-pair contrastive loss~\cite{sohn2016improved}:
\begin{equation}
    \loss_{\text{CTR}} = - \sum_{s, x} \log \frac{\exp(sim (u,v)/\tau)}{\sum_{x_j\in \mathcal{A}} \exp(sim(u, v(x_j)) /\tau)}
    \label{eq:contrastive}
\end{equation}
where $\mathcal{A} = \{\vecx\} \union \{\vecx_i^-\}_{i=1}^{N-1}$, 
$\tau$ is the temperature hyper-parameter, and
$sim$ is the cosine similarity function $sim(a,b)=a^\top b/\|a\|\|b\|$.
In the implementation, negative examples $\{\vecx_i^-\}_{i=1}^{N-1}$ are from the same training batch of data (Figure~\ref{fig:framework}(b)).

%% file: 024_data_aug.tex
\begin{figure}
    \setlength{\belowcaptionskip}{-0.4cm}
    \centering
    \includegraphics[width=\linewidth]{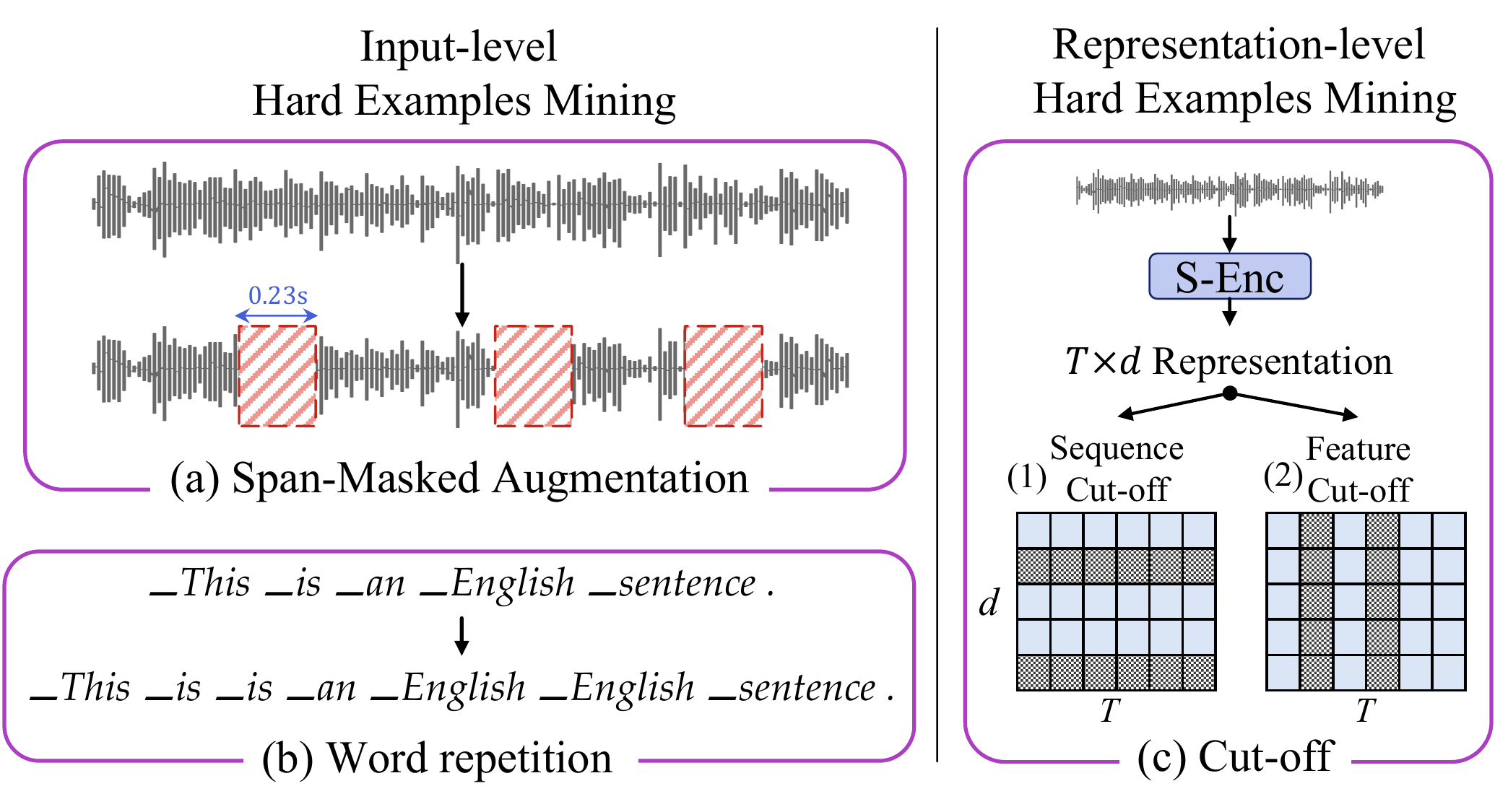}
    \caption{Schematic illustration of the hard examples mining strategies. In the cut-off strategy, the gray shaded grid represents the zero-out element.}
    \label{fig:da_illustration}
\end{figure}

To further enhance the contrastive learning, we introduce three strategies to mine additional hard examples. These strategies are at input and representation (gray shaded modules in Figure~\ref{fig:framework}(a)). Specific schematic illustrations of each operations are shown in Figure~\ref{fig:da_illustration}.

\noindent\textbf{Span-Masked Augmentation}~
We mask consecutive segments of an original audio waveform sequence $\vecs$ to obtain a new modified speech $\vecs'$. We take $\vecs'$ as an input to the model, and compute the contrastive loss on its original corresponding transcript.
We randomly sample without replacement all time steps in the original waveform of the speech to be the starting indices with a probability $p$, and then we set the sub-sequence $M$ successive time steps to be blank. In the experiment, we tried multiple configurations, and found $p=0.25, M=3600$ the best, resulting in a masked span of $0.225$ second. 
Since the masked speech fragment is very short, 
we consider the masked speech and the original transcript to be positive pairs, and the remaining transcripts in the same batch to be negative pairs.

\noindent\textbf{Word Repetition}
The word repetition strategy randomly replicates some words (or sub-words) in the original sentences, with two advantages for improving representation robustness.
First, as the length of the sentence is shorter than that of its audio representation, randomly repeating the words in the sentence is a simple yet useful technique to increase the length. 
Second, repeating words does not change the semantics and is suitable as an extra positive example of the corresponding speech.
Specifically, given sentence $\vecx$, each sub-word token $x_i$ can be duplicated \textbf{$k$ more} times, resulting in the duplicated sentence $\vecx'$, where $k=0,1,2,...$ and $k \sim \text{Poisson}(1)$.
We regard $\vecx'$ as the additional positive example for the speech $\vecs$ and the samples with the same operation in the same batch as the negative examples.

\noindent\textbf{Cut-off strategy}~
Recent studies on natural language understanding and generation have proved cut-off strategy to be successful~\cite{shen2020simple, yan2021consert}. 
We analogize a similar idea to the cut-off approach for speech representation. We entirely erase a slice of the $T \times d$ representation matrix along each dimension and set the erased terms to 0. Here, we present two variants: \textbf{\textit{sequence cut-off}}, which erases some sequence dimension, and \textbf{\textit{feature cut-off}}, which erases some feature dimension.
Note that there is a difference between cut-off and dropout. Dropout randomly sets some elements to 0, while cut-off is a dimensional ``block" dropout.
Similarly, we treat the cut-off audio representation and the original transcribed sentence as positive pairs, and the rest sentences in the same batch as negative pairs.

%% file: 030_exp.tex
\subsection{Experimental Setups}
\input{031_datasets}

\input{table_mustc_vs_baselines}

\input{032_setups}

\subsection{Main Results}
\input{034_result}

\input{table_vs_cascade}

%% file: 031_datasets.tex
\noindent\textbf{ST datasets}~
We conduct experiments on all the translation directions in \textbf{MuST-C} dataset~\footnote{We use v1.0. \url{https://ict.fbk.eu/must-c/}}~\cite{digangi2019must}: English (En) to German (De), Spanish (Es), French (Fr), Italian (It), Dutch (Nl), Portuguese (Pt), Romanian (Ro) and Russian (Ru).
As one of the largest ST benchmarks, MuST-C contains more than 385 hours of TED talks for each direction.

\noindent\textbf{MT datasets}~
We also introduce external \textbf{WMT} datasets~\cite{bojar2016findings} for En-De/Es/Fr/Ro/Ru and \textbf{OPUS100} datasets~\cite{zhang2020improving} for En-It/Nl/Pt directions, as the expanded setup.

Table~\ref{tab:data} (in Appendix.~\ref{sec:app_data}) lists the statistics of all the datasets included.

%% file: table_mustc_vs_baselines.tex
\begin{table*}[t]
\centering
\small
\resizebox{\textwidth}{!}{
\begin{tabular}{l|cccc|cccccccc|c}
    \toprule
    \multirow{2}{*}{\textbf{Models}} & \multicolumn{4}{c|}{\textbf{External Data}} & \multicolumn{9}{c}{\textbf{BLEU}} \\
    & \textbf{Speech} & \textbf{Text} & \textbf{ASR}  & \textbf{MT}  & 
    \textbf{De} & \textbf{Es} & \textbf{Fr} & \textbf{It} & \textbf{Nl} & \textbf{Pt} & \textbf{Ro} & \textbf{Ru} & \textbf{Avg.} \\ 
    \midrule
    \multicolumn{14}{c}{\emph{w/o external MT data}} \\
    \midrule
    Fairseq ST~\citep{wang2020fairseqs2t}  & - & - & - & - & 
        22.7 & 27.2 & 32.9 & 22.7 & 27.3 & 28.1 & 21.9 & 15.3 & 24.8 \\
    NeurST~\cite{zhao2021neurst} & - & - & - & - & 
        22.8 & 27.4 & 33.3 & 22.9 & 27.2 & 28.7 & 22.2 & 15.1 & 24.9 \\
    Espnet ST~\cite{inaguma2020espnet} & - & - & - & - & 
        22.9 & 28.0 & 32.8 & 23.8 &27.4 & 28.0 & 21.9 & 15.6 & 25.1 \\
    Dual Decoder~\cite{le2020dual} & - & - & - & - & 
        23.6 &28.1 & 33.5 & 24.2 & 27.6 & 30.0 & 22.9 & 15.2 & 25.6 \\
    W-Transf.~\cite{ye2021end} & \checkmark & - & - & - & 
        23.6 & 28.4 & 34.6 & 24.0 & 29.0 & 29.6 & 22.4 & 14.4 & 25.7 \\
    Speechformer~\cite{papi2021speechformer}& - & - & - & - &
        23.6 & 28.5 & - & -&  27.7 & - & - & - & - \\
    Self-training~\citep{pino2020self} & \checkmark & - &  \checkmark & - & 
        25.2 & - & 34.5 & - & - & - & - & -  & -\\
    SATE~\citep{xu2021stacked} & - & - & - & - & 
        25.2 & - & - & - & - & - & - & -  & -\\ 
    BiKD~\citep{inaguma2021source} & - & - & - & - & 
        25.3 & - & 35.3 & - & - & - & - & -  & - \\
    Mutual-learning~\cite{zhao2021mutual} & - & - & - & - & 
        - & 28.7 & 36.3 & -& - &-&-& - & - \\
    XSTNet~\cite{ye2021end} & \checkmark & - & - & - & 
        25.5 & 29.6 & 36.0 & 25.5 &	30.0 &31.3 & \textbf{25.1} & 16.9 & 27.5 \\
    \textbf{\method} & \checkmark & - & - & - & 
        \textbf{25.7} & \textbf{30.4} & \textbf{36.8} & \textbf{26.3} & \textbf{30.6} & \textbf{32.0} & 24.8 & \textbf{17.3} & \textbf{28.0} \\
    
    \midrule
    \multicolumn{14}{c}{\emph{w/ external MT data}} \\
    \midrule
    MTL~\cite{tang2021general}  & - & - & - & \checkmark  & 
        23.9 & 28.6 & 33.1 & - & - & - & - & - & -\\
    LightweightAdaptor~\cite{le2021lightweight} & \checkmark & \checkmark & -  & \checkmark  & 
        24.7 & 28.7 & 35.0 & 25.0 &  28.8 & 31.1 & 23.8 & 16.4 & 26.6 \\
    FAT-ST (Big)~\citep{zheng2021fused}  & \checkmark  &\checkmark  & \checkmark  & \checkmark  & 
        25.5 & 30.8 & - & - & 30.1 & - & - & - & - \\
    JT-S-MT~\cite{tang2021improving} & - & - & - & \checkmark &
        26.8 & 31.0 & 37.4 &- &-&-&-& - & - \\
    Chimera~\cite{han2021learning}  & \checkmark & - & -  & \checkmark  & 
        27.1\textsuperscript{\textdagger} & 30.6 & 35.6 & 25.0 & 29.2 & 30.2 & 24.0 & 17.4  & 27.4 \\
    XSTNet~\cite{ye2021end} & \checkmark & - & -  & \checkmark  & 
        27.1 & 30.8 & 38.0 & 26.4 & 31.2 & 32.4 & \textbf{25.7} & 18.5 & 28.8 \\
    SATE~\cite{xu2021stacked}  & - & - & \checkmark  & \checkmark  & 
        28.1\textsuperscript{\textdagger}  & - & -&-&-&-& -  & - & -\\ 
    STEMM~\cite{fang2022stemm} & \checkmark & - & -  & \checkmark  & 
        28.7 & 31.0 & 37.4 & 25.8 & 30.5 & 31.7 & 24.5 & 17.8 & 28.4 \\
    TaskAware~\cite{indurthi2021task} & - & - & \checkmark  & \checkmark  &
        \textbf{28.9}& - &- &-&-&-& - & - & - \\
    STPT~\cite{tang2022unified} & \checkmark & \checkmark & \checkmark  & \checkmark  &
        - & \textbf{33.1} & \textbf{39.7} &-&-&-&-& - & -\\
    \textbf{\method}  & \checkmark & - & -  & \checkmark  &
        28.3  & 32.0 & 38.3 & \textbf{27.2} & \textbf{31.7} & \textbf{33.1} & 25.6 & \textbf{18.9} & \textbf{29.4} \\ 
    \bottomrule
\end{tabular}
}
\caption{Case-sensitive detokenized BLEU scores on MuST-C \texttt{tst-COMMON} set. "Speech" denotes unlabeled speech data. "Text" means unlabeled text data, \textit{e.g.} Europarl V7~\cite{koehn2005europarl}, CC25~\cite{liu2020multilingual}. \textdagger \ use external 40M OpenSubtitles~\citep{lison2016opensubtitles2016} MT data. Other models only use WMT data.
}
\label{tab:main_result_vs_e2e}
\end{table*}

%% file: 032_setups.tex
\noindent\textbf{Model Configurations}~
The Wav2vec2.0 in the S-Enc is only pre-trained on Librispeech~\cite{panayotov2015librispeech} speech without any downstream fine-tuning\footnote{\url{https://dl.fbaipublicfiles.com/fairseq/wav2vec/wav2vec_small.pt}}.
Two layers of CNNs after the Wav2vec2.0 are set to kernel size 5, stride size 2 and hidden size 512.
The Transformer follows the base configuration, with 6 layers of encoder and decoder, hidden size $d=512$, 8 attention heads, and 2048 FFN hidden states. We use pre-layer normalization for stable training.
The model with the above configurations has a total of about 150M parameters.

\noindent\textbf{Experiment Details}~
We evaluate case-sensitive detokenized BLEU using sacreBLEU\footnote{\url{https://github.com/mjpost/sacrebleu}, \textbf{BLEU Signature}:
nrefs:1 | bs:1000 | seed:12345 | case:mixed | eff:no | tok:13a | smooth:exp | version:2.0.0}~\cite{post2018call} on MuST-C \texttt{tst-COMMON} set. In the analysis,  we also report the ChrF++ score~\footnote{\textbf{ChrF2++ Signature}: nrefs:1 | bs:1000 | seed:12345 | case:mixed | eff:yes | nc:6 | nw:2 | space:no | version:2.0.0}~\cite{popovic2017chrf++} and the learning-based BLEURT score~\footnote{\url{https://github.com/google-research/bleurt}~\cite{sellam2020bleurt}. As recommended, the checkpoint we use is \texttt{BLEURT-20}.}.
We use the raw 16-bit 16kHz mono-channel \textit{speech} input. We jointly tokenize the bilingual \textit{text} using SentencePiece~\cite{kudo2018sentencepiece}, with a vocabulary size of 10k, which is the same as \citet{ye2021end}'s setup.
For the training loss, we set contrastive temperature $\tau=0.02$ and weight of contrastive term $\lambda=1.5$ for German and Dutch, and $\lambda=1.0$ for the other languages. 

Appendix~\ref{sec:app_setup} contains more detailed settings and explanations for the baseline models in Table~\ref{tab:main_result_vs_e2e}. Appendix~\ref{sec:app_hyperparam} shows the experiments on the choice of the hyper-parameters.

%% file: 034_result.tex
\noindent\textbf{Comparison with end-to-end ST models}~
Table~\ref{tab:main_result_vs_e2e} shows the main results.
Since many existing works regard ``leveraging external data'' to be one of their model's features, their strong performances are largely predicated on the utilization of auxiliary data, especially large-scale MT data. For a relatively fair comparison, we investigate two cases: (1) without external MT data and (2) with external MT data.
Without the external MT data, our method already gains an average improvement of 0.5 BLEU over the previous best models.
Also when speech data is introduced for pre-training, our method works better than others (Self-training, W-Transf. and XSTNet).
When extra MT data are introduced, our method also outperforms SOTA by an average of 0.6 BLEU.
Among the benchmark models, with the same goal of closing two modality gaps, Chimera~\cite{han2021learning} constructed an extra fixed-length shared semantic space. However, the shared memory with fixed size actually compromises the MT performance, while our contrastive learning approach is more straightforward and effective.

\noindent\textbf{Comparison with cascaded ST systems}~
We compare our method with several cascade baselines, where \citet{ye2021end} and \citet{xu2021stacked} provided two strong cascade systems trained using MuST-C and external ASR and MT data (LibriSpeech, WMT, and Opensubtitles). From Table~\ref{tab:vs_cascade}, we find that as an end-to-end model, \method can outperform these strong cascade models. In Appendix~\ref{sec:app_case}, we provide a case study to show such improvement.

%% file: table_vs_cascade.tex
\begin{table}[ht]
    \setlength{\belowcaptionskip}{-0.5cm}
    \centering
    \resizebox{\columnwidth}{!}{
    \begin{tabular}{l|ccc}
        \toprule
        Models & \textbf{En-De} & \textbf{En-Fr} & \textbf{En-Ru} \\
        \midrule
        \textbf{Cascaded} \\
        \quad Espnet\cite{inaguma2020espnet} & 23.6  & 33.8 &  16.4 \\
        \quad \cite{ye2021end} & 25.2 & 34.9 & 17.0 \\
        \quad \cite{xu2021stacked} & 28.1 & - & - \\
        \midrule
        \textbf{End-to-end} \\
        \quad \method & \textbf{28.3} & \textbf{38.3} & \textbf{18.9} \\
        \bottomrule
    \end{tabular}
    }
    \caption{\method versus the cascaded ST systems on MuST-C En-De/Fr/Ru test sets.  \citet{ye2021end} and \citet{xu2021stacked} are two strong cascaded models.
    }
    \label{tab:vs_cascade}
\end{table}

%% file: 040_analysis.tex
\subsection{Is contrastive loss effective?}
\input{041_ctr_ablation}

\subsection{Which layer to contrast on?}
\input{042_where_to_contrast}
\input{table_where_to_contrast}

\subsection{Is contrastive loss better than other losses?}
\input{043_vs_other_losses}
\input{table_ctr_vs_l2}

\subsection{Analysis on the hard example mining strategies}
\input{044_da_tricks}

%% file: 041_ctr_ablation.tex
With the same model architecture and the same pre-training + fine-tuning procedure, the main difference between \method and XSTNet~\cite{ye2021end} is whether we use the contrastive loss term during the fine-tuning or not.
Comparing the BLEU results of w/o and w/ external MT data situations in Table~\ref{tab:main_result_vs_e2e}, we find that \method further improves $0.5$ and $0.6$ BLEU scores in terms of eight translation directions on average, which proves the effectiveness of the cross-modal contrastive learning.
By gradually removing each losses in Eq.(~\ref{eq:total_loss}), Table~\ref{tab:ablation} shows the improvements bringing by the multi-task learning and the contrastive learning.
For En-De translation direction, contrastive learning can bring an average improvement of 0.9 BLEU over the baseline models by only optimizing $\loss_{\text{ST}}$ (corresponding to the last row of the Table~\ref{tab:ablation}), and multi-task learning can lead to a further improvement of about 1.2 BLEU on top of that.

\input{table_ablation}

%% file: table_ablation.tex
\begin{table}[t]
    \setlength{\belowcaptionskip}{-0.4cm}
    \centering
    \begin{tabular}{l|cc}
        \toprule
         & \multicolumn{2}{c}{\textbf{External MT}} \\
        \cline{2-3}
        \textbf{Config.} &  \textit{without}  & \textit{with} \\
        \midrule
        \method & 25.7 & 28.3 \\
        \quad $-\loss_{\text{ASR}} - \loss_{\text{MT}}$ & 24.6 & 27.0\\
        \quad $-\loss_{\text{ASR}} - \loss_{\text{MT}}-\loss_{\text{CTR}}$ & 23.6 & 26.3 \\
        \bottomrule
    \end{tabular}
    \caption{BLEU scores on MuST-C En-De \texttt{tst-COMMON} set by removing individual losses. We test the results under settings with and without the introduction of external MT data.}
    \label{tab:ablation}
\end{table}

%% file: 042_where_to_contrast.tex
\label{sec:analysis_where_ctr}
An intriguing question is which representations should be considered in the contrastive loss function. In the method part (Section~\ref{sec:method_CTR}), we use averaged audio representation $u$ for speech $\vecs$ (Eq.(\ref{eq:audio_emb})) and averaged lexical embedding $v$ for the transcript $\vecx$ (Eq.(\ref{eq:text_emb})), denoted as \textit{low-level repr.}.
Whereas inspired by a recent study in multilingual MT~\citep{pan2021contrastive}, we also provide an alternative contrastive loss as a comparison, whose speech and text features are average-pooled semantic representations derived from the Transformer encoder, denoted as \textit{high-level repr.}.

Table~\ref{tab:where_to_ctr} shows that contrastive learning using \textbf{the low-level representations (\underline{Line 1}) is better than using the high-level ones (\underline{Line 2})}. On the other hand, although the performance of \underline{Line 2} is relatively inferior, it still outperforms the multi-task model without the contrastive loss (\underline{Line 3}).
The detailed analysis of possible explanations will be shown in Section~\ref{sec:retrieval}.

%% file: table_where_to_contrast.tex
\begin{table}[t]
    \setlength{\belowcaptionskip}{-0.6cm}
    \centering
    \resizebox{\linewidth}{!}{
    \begin{tabular}{l|ccc}
        \toprule
        \textbf{Representations} & \textbf{BLEU} & \textbf{ChrF++}  & \textbf{BLEURT} \\
        \midrule
        \textit{low-level repr.} & \textbf{28.3}* & \textbf{53.2}* & \textbf{64.5} \\
        \textit{high-level repr.} & 27.5$^\dagger$ & 52.6$^\dagger$ & 63.6\\
        w/o contrative loss & 27.1 & 52.1 & 62.4 \\
        \bottomrule
    \end{tabular}
    }
    \caption{BLEU, ChrF++ and BLEURT (\%) on En-De test set. Different representations are tested.
    *: \method is significantly better than the other two baselines ($p<0.01$). 
    $\dagger$: the model is  significantly better the baseline model without contrastive loss ($p<0.05$).
    }
    \label{tab:where_to_ctr}
\end{table}

%% file: 043_vs_other_losses.tex
\label{sec:analysis_vs_other_losses}
Our goal for introducing the contrastive loss term (denoted as CTR Loss) is to close the distance between speech and text representations. Whereas, there are other options to achieve this goal, such as
L2 loss and CTC loss.
\begin{itemize}[itemsep=1pt,leftmargin=10pt,parsep=0pt, topsep=1pt]
    \item \textbf{L2 Loss}: Without introducing any negative samples, L2 loss directly reduces the Euclidean distance between the representations of two modalities by minimizing $\loss=\|u - v\|^{2}$. L2 loss can be viewed as an implementation based on the idea of knowledge distillation~\cite{heo2019comprehensive,dong2021listen}.
    \item \textbf{CTC Loss}: The connectionist temporal classification (CTC) loss~\cite{graves2006connectionist} is commonly used in speech-related tasks~\cite{xu2021stacked, dong2021listen}. Unlike contrastive loss that cares about the representation, CTC loss connects the two modalities by establishing speech-text alignment and maximizing $p(\vecx|\veca)=\sum_{\pi \in \Pi_{\vecs,\veca}} \prod_{t=1}^{T} p_t(\pi_t|\veca)$, where $\Pi_{\vecs,\veca}$ is the set of all valid alignments.
\end{itemize}

Compared to the other two ways of bridging the modality gap, L2 and CTC loss, is the contrastive loss term better? The answer is yes according to the results in Table~\ref{tab:ctr_vs_l2}.
Our explanation is that information on the negative samples benefits the contrastive loss, bringing the the distance between the speech and its corresponding transcription closer while pushing the distance to the irrelevant text farther.

%% file: table_ctr_vs_l2.tex
    

\begin{table}[ht]
    \centering
    \small
    \resizebox{\linewidth}{!}{
    \begin{tabular}{l|ccc}
        \toprule
        \textbf{Extra Loss} & \textbf{BLEU} & \textbf{ChrF++} & \textbf{BLEURT} \\
        \midrule
        CTR Loss & \textbf{28.3}* & \textbf{53.2} & \textbf{64.5} \\
        CTC Loss & 27.6$^\dagger$ & 53.0$^\dagger$ & 64.1 \\
        L2 Loss & 27.3 & 52.4 & 63.0 \\
        - & 27.1 & 52.1 & 62.4 \\
        \bottomrule
    \end{tabular}
    }
    \caption{BLEU, ChrF++ and TER (\%) on En-De test set under different loss terms other than the basic multi-task NLL loss. 
    *: \method is significantly ($p<0.01$) better than the other three alternatives.  
    $\dagger$: the improvement from CTC loss over the baseline without extra loss is significant ($p<0.01$).
    }
    \label{tab:ctr_vs_l2}
\end{table}

%% file: 044_da_tricks.tex
\label{sec:exp_DA}
\begin{figure}[t]
    \setlength{\belowcaptionskip}{-0.4cm}
    \centering
    \includegraphics[width=0.9\linewidth]{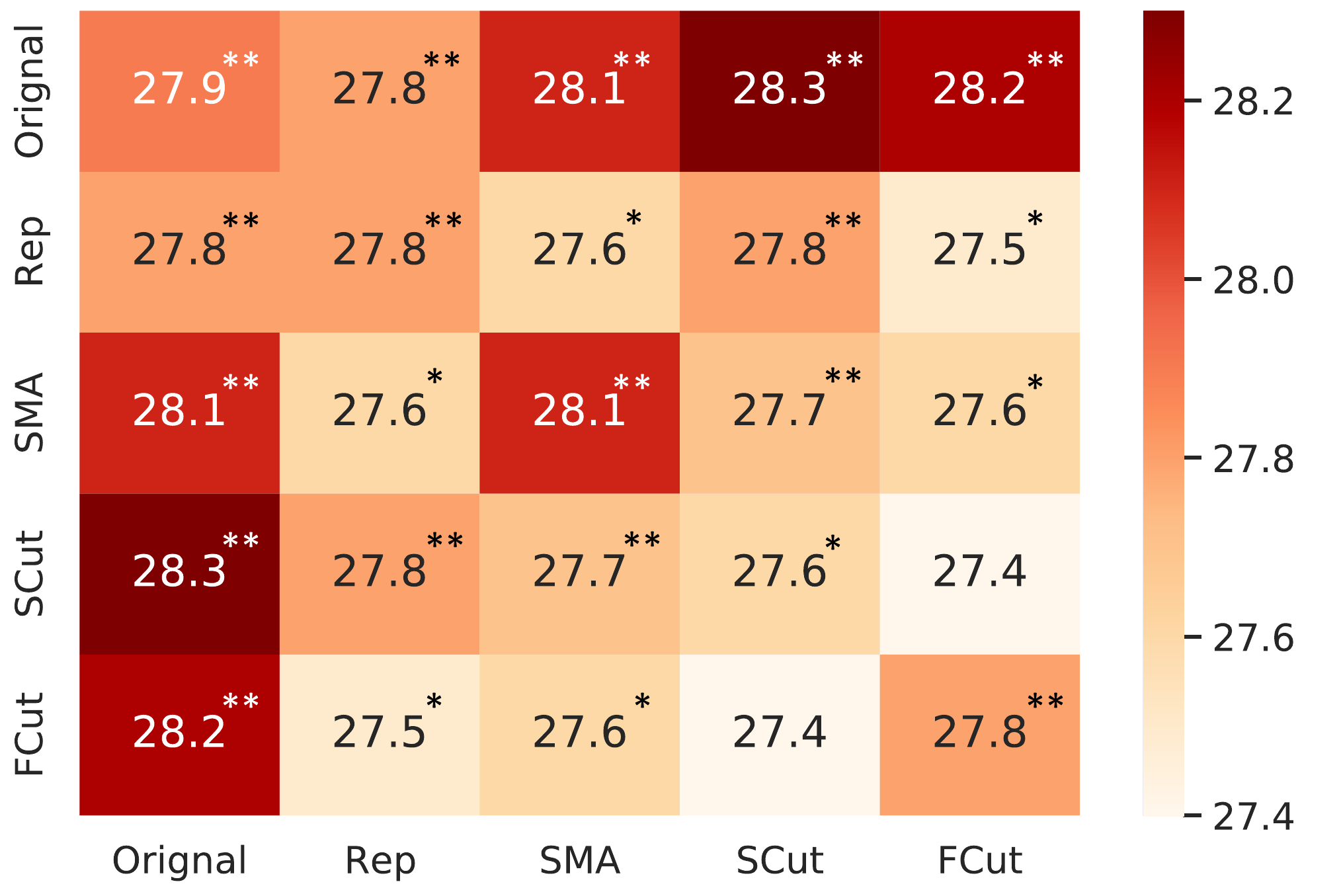}
    \caption{The heat map visualization of the BLEU scores on En-De test set, with 15 combinations of the original contrastive loss (Original) and hard examples mining methods -- word repetition (Rep), span-masked augmentation (SMA), sequence cut-off (SCut) and feature cut-off (FCut).
    * and ** mean the improvements over the XSTNet baseline without contrastive loss are statistically significant (*:$p<0.05$, **:$p<0.01$).}
    \label{fig:DA_method}
\end{figure}

In Section~\ref{sec:method_DA}, we proposed four methods to mine the hard examples for contrastive learning, namely span-masked augmentation (SMA), word repetition (Rep), sequence cut-off (SCut), and feature cut-off (FCut). In this section, we study how effective these methods are, and to do so, we consider the BLEU performances of their 15 combinations (Figure~\ref{fig:DA_method}). Note that ``Original'' means the original contrastive loss in Eq.(\ref{eq:contrastive}) without any additional hard examples mining operation, and the diagonal in the heat map represents only one strategy used. For an easy and fair comparison, we set the weight of the contrastive term to 1.0 uniformly.
We have the following observations.

\textbf{All the hard examples mining methods are effective.}~
All the BLEU scores in Figure~\ref{fig:DA_method} exceed the strong multi-task model trained without contrastive learning ($27.1$).
Among all the strategies, the combination of the original and SCut reaches the best result ($28.3$), and is better than the model without any expanded operations ($p<0.01$).
Generally, to find the best model, we suggest adopting multiple strategies and choosing the best checkpoint on the dev-set.

\textbf{The combinations of the hard examples mining methods and the ``original'' have relatively better performances.} We argue that we need the original positive and negative examples to give more accurate representations (without any dropout) for contrastive learning.
On the contrary, without the help of ``original'' loss, the performance with both sequence cut-off and feature cut-off is the worst in Figure~\ref{fig:DA_method}, probably because too much information is lost by superimposing the two.

%% file: table_case_study.tex
\begin{table*}[th]
    \setlength{\belowcaptionskip}{-0.4cm}
    \centering
    \small
    \begin{tabular}{l|cp{13cm}}
        \toprule
        \textbf{Models} &  &  \\
        \midrule
        \multicolumn{3}{c}{\textbf{CASE 1} } \\
        \midrule
        \multirow{2}{*}{Ref.}
         & \textbf{src}: & Lights, sounds, solar panels, motors — everything should be accessible. \\
         & \textbf{tgt}: & Lichter, Töne, Solarelemente, Motoren — alles sollte verfügbar sein. \\
        \midrule
        \multirow{2}{*}{Cascaded}
        & \textbf{src}: & 	Lights sounds solar panels motors everything should be accessible. \\
        & \textbf{tgt}: & Licht \textcolor{red}{\underline{klingt}} Solarpaneele, Motoren; alles sollte zugänglich sein. \\
        \midrule
        XSTNet & \textbf{tgt}: & Licht, Geräusche, Solarkollektoren, Motoren — alles sollte zugänglich sein.\\
        \midrule
        \method & \textbf{tgt}: &  Licht, Geräusche, Solarpanele, Motoren, alles sollte zugänglich sein. \\
        \midrule
        
        \multicolumn{3}{c}{\textbf{CASE 2} } \\
        \midrule
        \multirow{4}{*}{Ref.}
        & \textbf{src}: & Eight years ago when I was at the Media Lab, I started exploring this idea of how to put the power of engineers in the hands of artists and designers. \\
        & \textbf{tgt}: & Vor acht Jahren war ich am Media Lab und ich begann diese Idee zu erforschen, wie man die Macht der Ingenieure in die Hand von Künstlern und Designern legen könnte. \\
        \midrule
        \multirow{4}{*}{Cascaded}
        & \textbf{src}: & Eight years ago when I was at the Media Lab, I started exploring this idea of how to put the power of engineers in the hands of artists and designers. \\
        & \textbf{tgt}: & Vor 8 Jahren, als ich im Media Lab war, begann ich, diese Idee zu erforschen, wie man die Macht der Ingenieure in die Hände von Künstlern und Designern legte. \\
        \midrule
        \multirow{2}{*}{XSTNet}
        & \textbf{tgt}: & Vor acht Jahren, als ich im Media Lab war, \textcolor{red}{\underline{begann ich zu erforschen}}, wie man die Kraft der Ingenieure in die Hände von Künstlern und Designern legt. \\
        \midrule
        \multirow{2}{*}{\method}
        & \textbf{tgt}: & Vor acht Jahren, als ich im Media Lab war, begann ich, diese Idee zu erforschen, wie man die Macht von Ingenieuren in die Hände von Künstlern und Designern legt. \\
        \bottomrule
    \end{tabular}
    \caption{En-De test cases that generated from the cascaded model, XSTNet (both provided by~\citet{ye2021end}) and our \method model.
    The \textcolor{red}{\underline{red underlined text}} indicates grammatically incorrect or inaccurate translations.}
    \label{tab:case_study}
\end{table*}

%% file: 050_analysis_gap.tex
\label{sec:ana_gap}

As mentioned earlier, the existing multi-task training models cannot address the \textit{speech-text modality gap}. Does \method reduce the representation gap between speech and text?

\subsection{Visualization of Representation}
\input{052_visualization}

\subsection{Cross-modal Retrieval}
\input{053_retrieval}
\input{table_retrieval_acc}

%% file: 052_visualization.tex
\begin{figure}
    \setlength{\belowcaptionskip}{-0.6cm}
    \centering
    \subfigure[w/o contrastive loss]{
        \begin{minipage}{0.46\linewidth}
            \centering
            \includegraphics[scale=0.25]{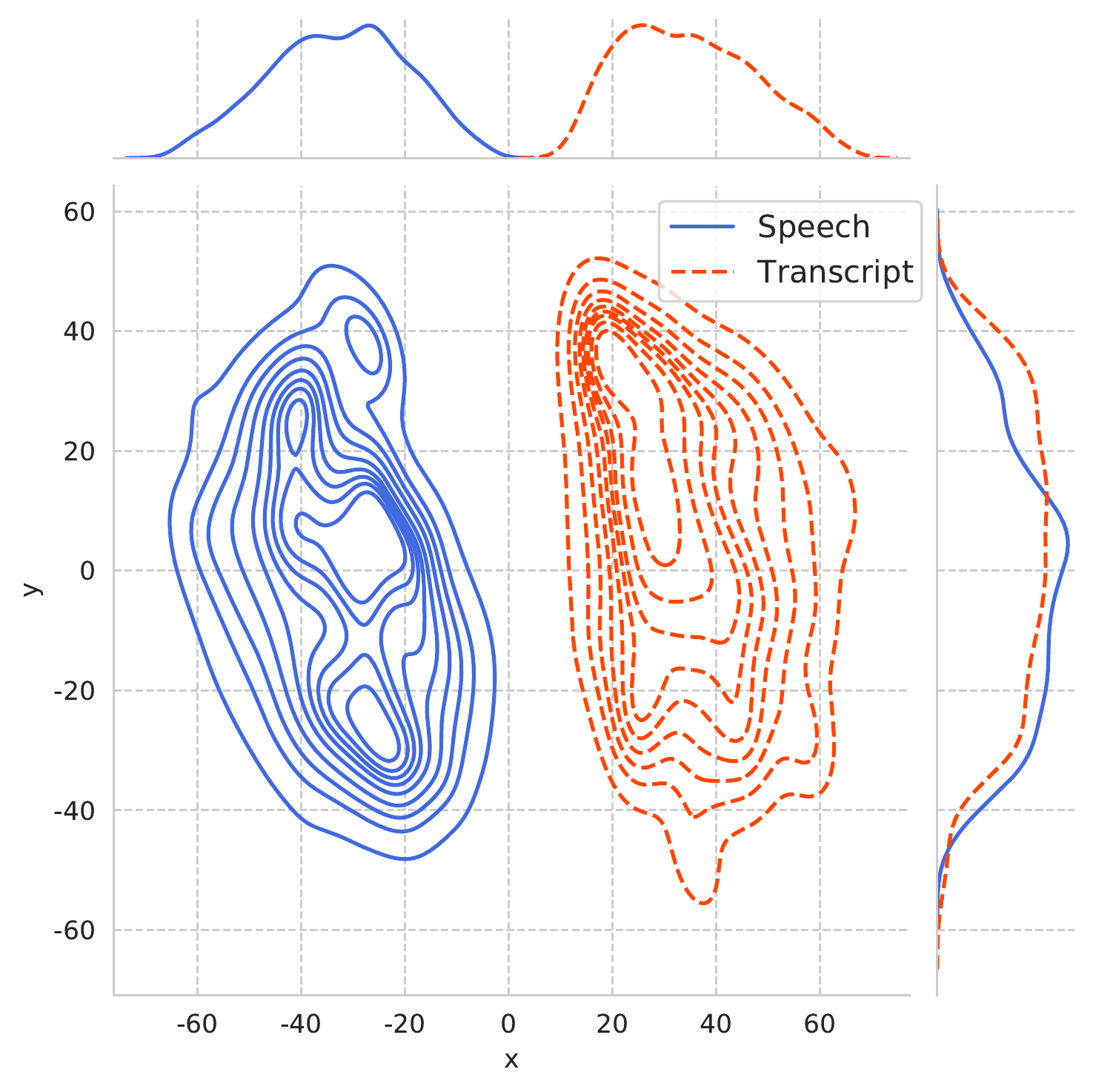}
        \end{minipage}
    }
    \subfigure[\method]{
        \begin{minipage}{0.46\linewidth}
            \centering
            \includegraphics[scale=0.25]{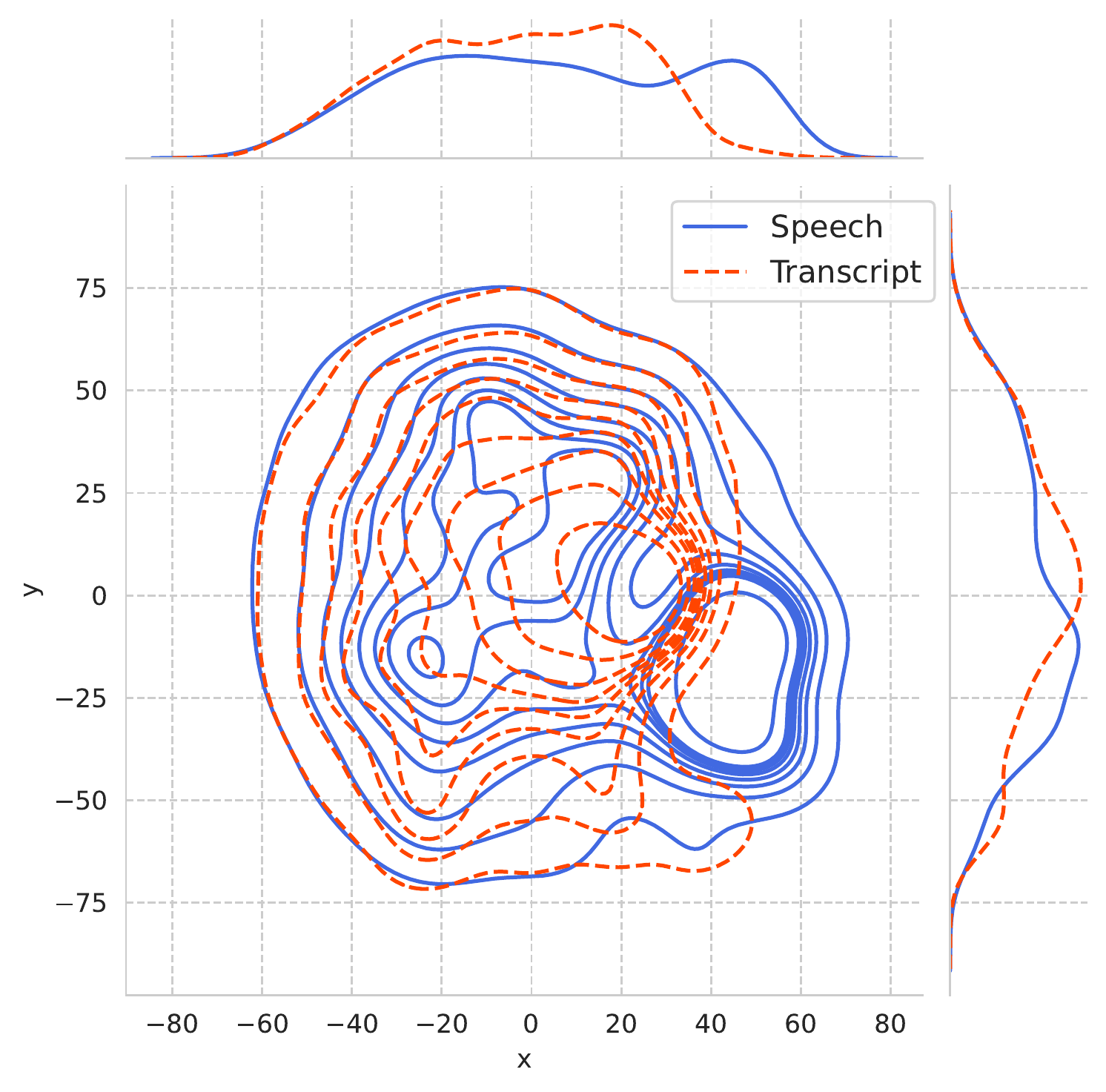}
        \end{minipage}
    }
    \caption{
    Bivariate KDE contour plot of the representation of speech and transcript in source language English. T-SNE is used to reduce into 2D. The blue lines are the audio representations and the red dashed lines stand for text. (a) for the vanilla multi-task framework without any extra supervision. (b) for our proposed \method model. Sentences are from En-De test set.}
    \label{fig:kde_analysis}
\end{figure}

\noindent\textbf{Does the \textit{speech-text modality gap} exist without explicitly bridging the two?}
\emph{Speech-text modality gap} means the discrepancy between the audio representations  and transcription sentence embeddings. To visualize it, we plot the bivariate kernel density estimation~\cite{parzen1962estimation}~(KDE) contour of their dim-reduced features, where T-SNE~\citep{van2008visualizing} is used to reduce the dimension into two (Figure~\ref{fig:kde_analysis}).
Ideally, if the representations of speech and its corresponding transcript are similar, their KDEs will be similar, and thus the contour lines will overlap as much as possible.
However, Figure~\ref{fig:kde_analysis}(a) is the KDE contour of the multi-task framework without any explicit modeling to bring two modalities together~\cite{ye2021end}. It shows that the representations are so dissimilar that they are organically divided into two clusters, \textit{i.e.} \textit{speech-text modality gap} exists.

\noindent\textbf{Does \method reduce the modality gap?}
As shown in Figure~\ref{fig:kde_analysis}(b), compared to the baseline model without contrastive learning, \method with cross-modal contrastive learning is able to bring representations of different modalities much closer.
This means that the audio representation contains more linguistic information similar to that of the textual transcription, which is more advantageous for the downstream ST generation through the shared \textit{Transformer encoder} and \textit{Transformer decoder}.

%% file: 053_retrieval.tex
\label{sec:retrieval}
\textbf{How good is the cross-modal representation space learned from \method?}
To answer this question, we conduct a retrieval experiment, \textit{i.e.} finding the nearest (smallest cosine similarity) transcript based on the speech representation.
We compare \method model with the baseline without cross-modal contrastive learning and report the top-1 retrieval accuracy using (1) the low-level representations and (2) the high-level semantic representations, in Table~\ref{tab:retrieval_acc}.

When retrieving the text using low-level representations, our method gains a substantial 79\% increase over the baseline. 
In addition, we find that without explicit contrastive modeling, the baseline can achieve retrieval accuracy of more than $94\%$ according to the semantic representations outputted from the \textit{Transformer encoder}. We believe that such high accuracy is automatically learned from the triple-supervised data itself under the multi-task learning framework.
With such a degree of cross-modal alignment, if we construct the contrastive loss with semantic representations, its gain to the ST performance turns out to be limited, which exactly corroborates the findings in Section~\ref{sec:analysis_where_ctr} -- low-level representations are preferred in the cross-modal contrastive learning.

%% file: table_retrieval_acc.tex
\begin{table}[ht]
    \setlength{\belowcaptionskip}{-0.6cm}
    \centering
    \begin{tabular}{l|c|r}
        \toprule
        \textbf{Representations} & \textbf{CTR loss} & \textbf{Acc.} \\
        \midrule
        \multirow{2}*{low-level repr.} & \texttimes & 9.4 \\
        & \checkmark & \textbf{88.6} \\
        \midrule
        \multirow{2}*{high-level repr.} & \texttimes & 94.7 \\
        & \checkmark & \textbf{95.0} \\
        \bottomrule
    \end{tabular}
    \caption{Cross-modal top-1 retrieval accuracy~(\%) on En-De test set. Two different representations are used, based on which, \method achieves huge accuracy improvements.}
    \label{tab:retrieval_acc}
\end{table}

%% file: 093_case_study.tex
\label{sec:app_case}

In this section, we use several cases that \method generates. We compare our model with the cascaded model and the previous end-to-end model, XSTNet~\cite{ye2021end}.

For this first case, the cascaded system fails to give a right translation due to the mis-punctuation issue (\textit{klingt} is a verb), while the end-to-end model, XSTNet and \method translate correctly.
For the second case, the previous end-to-end XSTNet model cannot accurately translate the phrase ``started exploring this idea of'', which performs worse than the cascaded one. Whereas \method successfully conveys the meaning of ``this idea'' , and translates more accurately than XSTNet.
We believe this improvement comes from the cross-modal contrastive learning.

%% file: 070_conclusion.tex
In this paper, we propose \method, a simple yet effective contrastive learning framework bridging the speech-text representation gap and facilitating the ST with limited data. We also provide feasible hard example mining methods to learn robust representations. 
The results on the MuST-C ST dataset prove the effectiveness of the method.

%% file: 080_broader_impact.tex
This work improves the performance of ST tasks on public datasets by learning speech representations that are more similar to text representations, but the model is far from being achieved for industrial-grade implementations.
In real scenarios, for example, the original voice is noisier and the distribution of speech lengths is more complex than in the public dataset, which cannot be handled by an end-to-end model alone.
The shortcoming of this model is that it still needs a certain amount of labeled data for training, especially \textit{<speech,transcription>} to learn better speech representation, and for the more than $7,000$ languages and dialects in the world, most of them do not have corresponding translations or even transcriptions, our method does not work in untranscribed scenarios.
In this paper, we focus on the improvement brought by the better speech representation on the ST task, and obtained good results with hundreds of hours of speech data. We hope that our work achieves better results using more data (\textit{e.g.} raw speech, raw text, ASR, MT data) in the future.

%% file: 090_appendix.tex
\section{Statistics of all datasets}
\label{sec:app_data}
\input{table_data}

\section{Experimental Details}
\label{sec:app_setup}
\noindent\textbf{Training and Implementation Details}~
We use Adam optimizer ($\beta_1=0.9, \beta_2=0.98$) with learning rate $=1e^{-4}$ and warmup 25k steps during the ST training.
We also implement the expanded setting with the introduction of external WMT to train the Transformer module. In the pre-training stage, we set the learning rate $=7e^{-4}$ and warmup 4000 steps.
For robust training, we set label smoothing to $0.1$, and dropout rate to $0.1$.  
The hyper-parameters for different data augmentation methods are as follows: for masked audio span strategy, we set masking probability $p=0.25$ and masking span length $M=3600$ frames; for both sequence and feature cut-off, we set the cut-off dropout rate as $0.1$.
We save the checkpoint with the best BLEU on dev-set and average the last 10 checkpoints. For decoding, we use a beam size of 10 and length penalty $0.7$ for German, $1.0$ for French, and $0.4$ for Russian.
We train the models in 8 Nvidia Tesla V100 GPUs for each experiment.
We use Fairseq~\cite{ott2019fairseq} as the code-base for our implementation.

\noindent\textbf{Baseline Models}~
\input{033_baseline}

\section{The Choice for Hyper-parameters}
\label{sec:app_hyperparam}
\input{091_ablation_hyperparam}

\section{Data Scale for Fine-tuning}
\input{092_datascale}

%% file: table_data.tex
\begin{table}[h]
    \centering
    \begin{tabular}{c|cr||cr}
        \toprule
         & \multicolumn{2}{c||}{\textbf{ST (MuST-C)}} & \multicolumn{2}{c}{\textbf{MT}} \\
         \textbf{En$\rightarrow$} & hours & \#sents & name & \#sents \\
        \midrule
         \textbf{De} & 408 & 234K & WMT16 & 4.6M \\
         \textbf{Es} & 504 & 270K & WMT13 & 15.2M\\
         \textbf{Fr} & 492 & 292K & WMT14 & 40.8M \\
         \textbf{It} & 465 & 258K & OPUS100 & 1.0M\\
         \textbf{Nl} & 442 & 253K & OPUS100 & 1.0M \\
         \textbf{Pt} & 385 & 211K & OPUS100 & 1.0M \\
         \textbf{Ro} & 432 & 240K & WMT16 & 0.6M \\
         \textbf{Ru} & 489 & 270K & WMT16 & 2.5M \\
        \bottomrule
    \end{tabular}
    \caption{Statistics of all datasets}
    \label{tab:data}
\end{table}

%% file: 033_baseline.tex
In Table~\ref{tab:main_result_vs_e2e}, we compared our method with end-to-end baseline models whose audio inputs are 80-channel log Mel-filter bank, including:
FairseqST~\cite{wang2020fairseqs2t}, NeurST~\cite{zhao2021neurst}, Espnet ST~\cite{inaguma2020espnet},
Dual-decoder Transformer~\cite{le2020dual}, 
SATE~\cite{xu2021stacked}, 
Speechformer~\cite{papi2021speechformer},
self training~\cite{pino2020self} and mutual learning~\cite{zhao2021mutual} method,
STAST~\cite{liu2020bridging},
bi-KD~\cite{inaguma2021source},
MLT method~\cite{tang2021general},
Lightweight Adaptor~\cite{le2021lightweight},
JT-S-MT~\cite{tang2021improving},
FAT-ST~\cite{zheng2021fused},
TaskAware~\cite{indurthi2021task},
and STPT~\cite{tang2022unified}.
We also compare our method to baseline models that have pretrained Wav2vec2.0 as a module, including:
\begin{itemize}[itemsep=1pt, leftmargin=10pt, parsep=0pt, topsep=1pt]
    \item \textbf{W-Transf.}~\cite{ye2021end}: the model has the same structure as ours, but is only trained on \emph{<speech, translation>} parallel data.
    \item \textbf{Chimera-ST}~\cite{han2021learning}: the model that builds a shared semantic memory for both audio and text modalities.
    \item \textbf{XSTNet}~\cite{ye2021end}: the model has the same structure as ours, and adopted a multi-task fine-tuning strategy.
    \item \textbf{STEMM}~\cite{fang2022stemm}: the model that bridges the modality representation gap by minimizing the Jensen–Shannon divergence between the original speech representation and the manifold mix-up representation. 
\end{itemize}

%% file: 091_ablation_hyperparam.tex
\begin{figure}[hbt]
    \centering
    \includegraphics[width=0.9\linewidth]{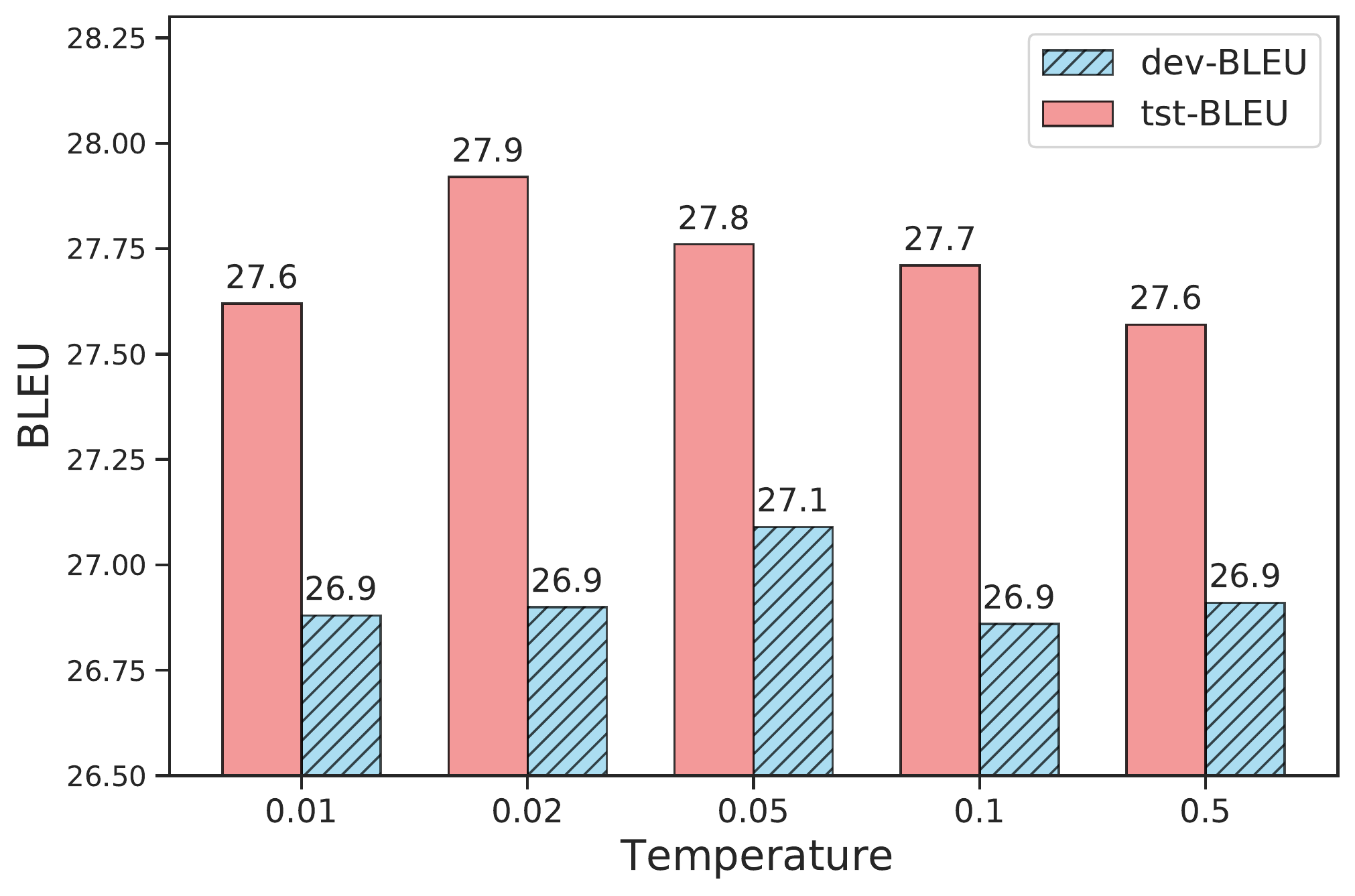}
    \caption{En-De BLEU scores on \texttt{tst-COMMON} and \texttt{Dev} set. the x-axis is the choices of different temperature $\tau$ in Eq.(\ref{eq:contrastive}) varying from $0.01$ to $0.5$.}
    \label{fig:temperature}
\end{figure}

\noindent\textbf{Influence of Temperature}~
In the contrastive loss, the temperature hyper-parameter is provided to control the smoothness of the distribution normalized by softmax operation.
A high temperature helps to smooth the distribution, making it more difficult for the model to distinguish between positive and negative samples (corresponding to correct transcriptions and other transcriptions in this work), while the low temperature behaves just the opposite.
We choose several temperature hyper-parameters ranging from $0.01$ to $0.5$, and Figure~\ref{fig:temperature} shows their BLEUs on the test and dev sets . We find that (1) the choice of the temperature does not drastically affect the final BLEU score, 
and (2) we recommend that the temperature $\tau$ be set between 0.02 and 0.05 to ensure a relatively good ST performance. In the experiment, we use $\tau=0.02$.

\begin{figure}[hbt]
    \centering
    \includegraphics[width=0.9\linewidth]{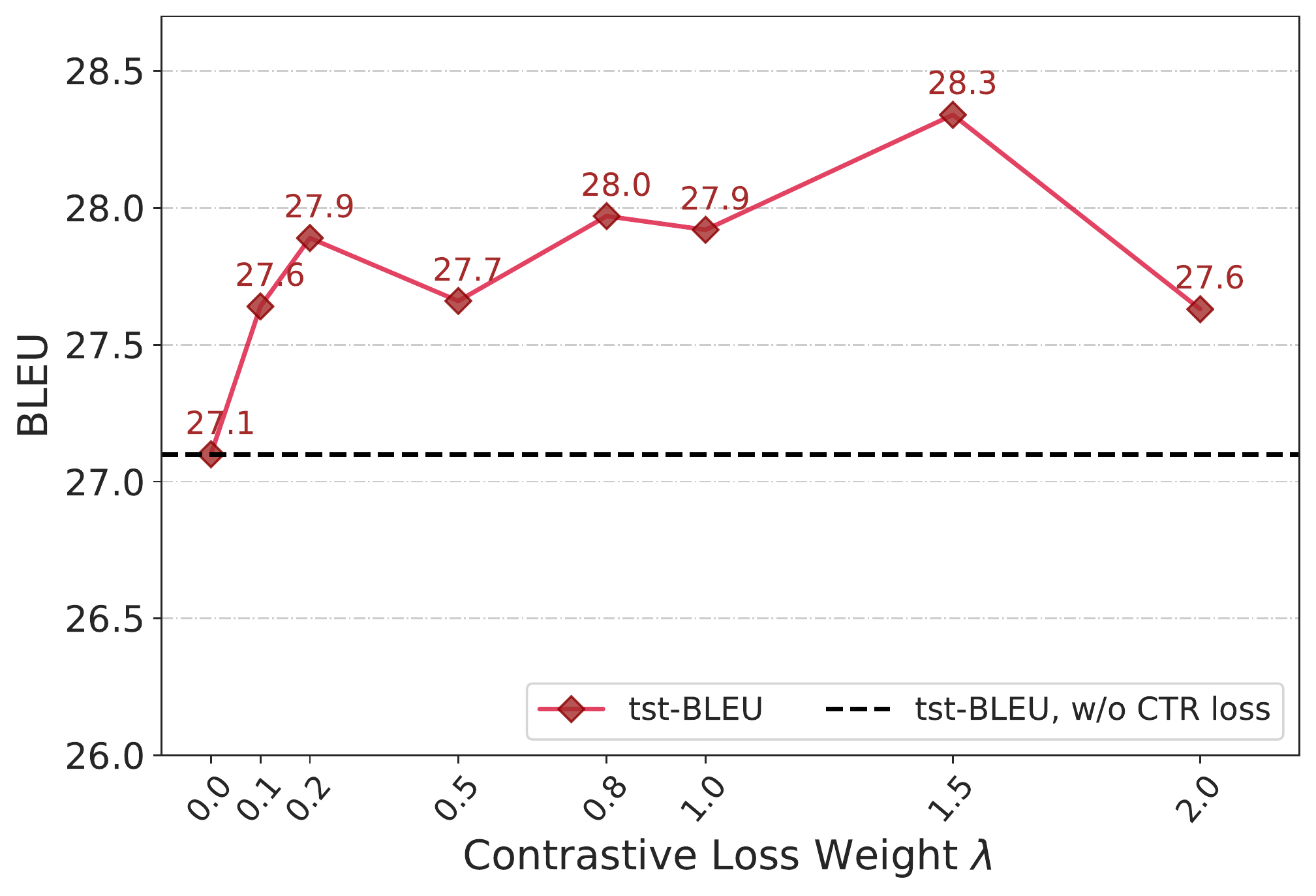}
    \caption{En-De BLEU scores on \texttt{tst-COMMON} and \texttt{Dev} sets. 
    The x-axis is the weight of the contrastive loss term $\lambda$ in Eq.(\ref{eq:total_loss}). Experiments are performed under the fix temperature hyper-parameter $\tau = 0.02$.
    }
    \label{fig:ctr_loss_weight}
\end{figure}

\noindent\textbf{Influence of Contrastive Loss Weight}~
The total loss we optimize, Eq.(\ref{eq:total_loss}), is a linear combination of the multi-task cross-entropy losses $\loss_{\text{MLT}}$ and the contrastive term $\loss_{\text{CTR}}$. 
To investigate how much the contrastive terms affect BLEU, we fix its temperature $\tau=0.02$, adjust the values of its loss weight $\lambda$ from 0.1 to 2.0, performed three experiments for each value, and test the average BLEU on En-De \texttt{tst-COMMON} set.
Figure~\ref{fig:ctr_loss_weight} depicts the performances.
First, all objective functions containing $\loss_{\text{CTR}}$, even if their weights $\lambda$ take different values, are apparently better than the baseline model with $\loss_{\text{MLT}}$ only $\loss_{\text{CTR}}$.
Then, the best BLEU score is achieved at loss weight $\lambda=1.5$, corresponding to the results in Table~\ref{tab:main_result_vs_e2e}. 
And when analyzing the effect of data augmentation strategies (Section~\ref{sec:exp_DA}), since we need to consider the combination between them, which is more complicated. Therefore, we set the loss weight to $1.0$ uniformly for simplicity.
In general, we recommend that the weight hyper-parameter takes a value between $0.8$ and $1.5$.


%% file: 092_datascale.tex
The experiments in the main paper show that our model can perform well by introducing external MT data pre-training. Here, we simulate the scenario with plenty of MT and speech data and limited ST triple-labeled data, and does \method have the ability of low-resource learning?
In the experiment, we reduce the labeled ST data to 1, 10, and 100 hours, corresponding to sentence counts of about 500, 5k, and 50k sentences. For a fair comparison, we use the same MT pre-trained Transformer module as in the main paper.
We find the contrastive loss particularly helpful when the amount of speech data is extremely small, like only 1 hour of speech. Second, the multi-task training strategy is also very effective in improving the robustness of the model performance.
We also find that by using easily accessible MT and speech pre-training, our model could reach the previous baseline results without pre-training using only $1/4$ of the original data, \textit{i.e.} $100$ hours of labeled ST data.

\begin{figure}[th]
    \centering
    \includegraphics[width=1.0\linewidth]{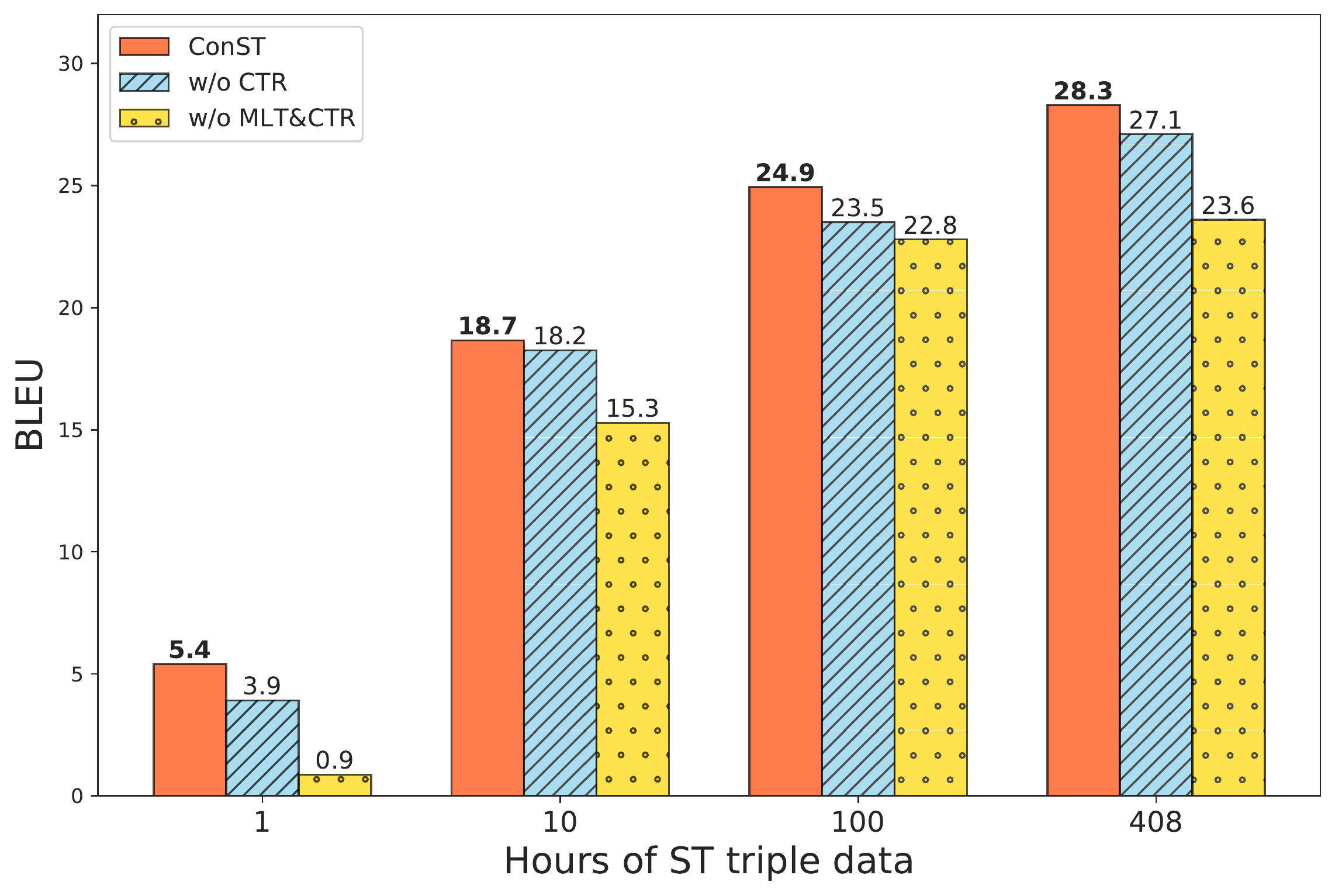}
    \caption{En-De BLEU scores on \texttt{tst-COMMON} sets. 
    The horizontal axis is the amount of ST data (in hours of speech).}
    \label{fig:datascale_exp}
\end{figure}